\pdfoutput=1

\documentclass[11pt]{article}

\usepackage[preprint]{acl}

\usepackage{times}
\usepackage{latexsym}

\usepackage[T1]{fontenc}
\usepackage[utf8]{inputenc}
\usepackage{microtype}
\usepackage{inconsolata}
\usepackage{graphicx}
\usepackage{amsmath}

\usepackage{amssymb}

\usepackage{algorithmic,algorithm}
\renewcommand{\algorithmiccomment}[1]{\bgroup\hfill$\triangleright$~#1\egroup}
\usepackage[linguistics]{forest} 
\usepackage{synttree}
\usepackage{subcaption}
\usepackage[cjk]{kotex}

\usepackage{booktabs}

\usepackage{amsthm}
\newtheorem{theorem}{Theorem}

\newenvironment{proofsketch}{\par\noindent\textit{Proof sketch}\ }{\hfill$\square$\par}

\usepackage{tikz-dependency}
\usepackage{algorithmic,algorithm}
\renewcommand{\algorithmiccomment}[1]{\bgroup\hfill$\triangleright$~#1\egroup}
\usepackage[linguistics]{forest} 

\usepackage{langsci-gb4e}

\title{A tree interpretation of arc-standard dependency derivations}

\author{
Zihao Huang\raisebox{-0.1\height}{\includegraphics[height=0.7em]{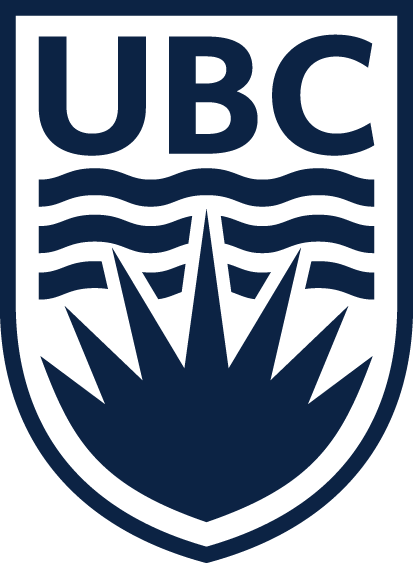}}~~~
Ai Ka Lee\raisebox{-0.1\height}{\includegraphics[height=0.7em]{UBC.png}}~~~
Jungyeul Park\raisebox{-0.1\height}{\includegraphics[height=0.7em]{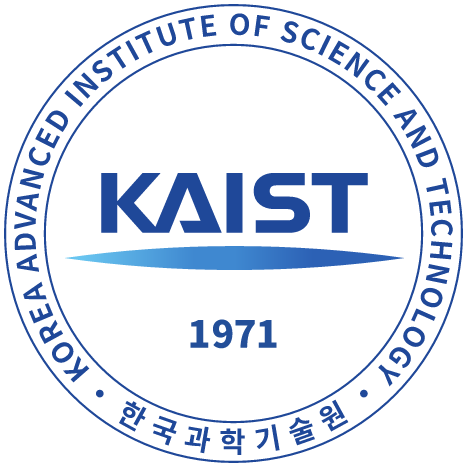}}\thanks{~~Corresponding author.}\\
\raisebox{-0.1\height}{\includegraphics[height=0.9em]{UBC.png}} The University of British Columbia, Vancouver, Canada\\
\raisebox{-0.1\height}{\includegraphics[height=0.9em]{KAIST}} Korea Advanced Institute of Science \& Technology, Daejeon, South Korea\\
\url{https://linguistics.ubc.ca}~~~~~~~~ \url{https://ct.kaist.ac.kr}
\\}

\begin{document}
\maketitle

\begin{abstract}
Arc-standard derivations over projective dependency trees can be interpreted as the incremental construction of lexicalized ordered trees with contiguous yields.
Each \textsc{shift}, \textsc{leftarc}, and \textsc{rightarc} transition corresponds to a deterministic tree update, and the resulting ordered tree uniquely determines the dependency arcs introduced by the derivation.
We show that this representation is not an arbitrary encoding: a single-headed dependency tree admits such a contiguous ordered representation if and only if it is projective.
The proposal is therefore derivational rather than conversion-based, since the ordered object is defined over the transition sequence itself rather than obtained by transforming a completed dependency graph.
This gives a tree-theoretic interpretation of arc-standard parsing, in which projective dependency derivations implicitly construct recoverable constituency-style ordered trees.
For non-projective inputs, the interpretation can be used through pseudo-projective lifting and inverse decoding.
A small implementation study confirms that the mapped derivations are executable in an existing neural transition-based parser.
\end{abstract}

\section{Introduction}

Dependency and constituency parsing encode syntactic structure through different
primitive objects. Dependency representations describe syntax as directed
head--dependent relations among lexical items
\citep{tesniere-1966-elements,melcuk-1988-dependency}, whereas constituency
representations describe syntax as hierarchically organized phrase-structure
trees \citep{chomsky-1957-syntactic}. Both traditions also admit
transition-based formulations over stack--buffer configurations: arc-based
dependency parsers incrementally construct dependency arcs
\citep{nivre-2003-efficient,yamada-matsumoto-2003-statistical,chen-manning-2014-fast},
while transition-based constituency parsers incrementally construct ordered
tree fragments
\citep{sagae-lavie-2005-classifier,zhu-etal-2013-fast,dyer-etal-2016-recurrent}.
What remains less explicit is whether these two derivational views are related
before the final syntactic object has been completed.

Prior work has related dependency and constituency mainly through completed
structures. Dependency analyses have been converted into phrase-structure
representations \citep{xia-palmer-2001-converting,kong-etal-2015-transforming},
and dependency trees have been encoded as bracketed, linearized, or ordered
outputs for linearization, surface realization, and sequence-based prediction
\citep{bohnet-etal-2012-generating,song-etal-2018-neural,puzikov-gurevych-2018-binlin,yu-etal-2020-fast}.
Most closely related, \citet{fernandez-gonzalez-martins-2015-parsing}
introduce head-ordered dependency trees as an enriched dependency-graph
representation for reducing constituency parsing to dependency parsing: headed
constituency trees are encoded as dependency graphs with attachment-order
information and are parsed by a dependency parser before constituency recovery.
This establishes an important connection between ordered dependency structure
and constituency parsing, but the ordered object is still a dependency graph
derived from completed headed constituency trees.

This paper proceeds in the opposite direction. We start from dependency trees
and arc-standard derivations, and show that projective derivations can be
interpreted as the incremental construction of lexicalized ordered trees whose
subtrees have contiguous yields. Each \textsc{shift}, \textsc{leftarc}, and
\textsc{rightarc} transition corresponds to a deterministic tree update, and
the resulting ordered tree uniquely determines the dependency arcs introduced
by the derivation. The proposed object is therefore not a head-ordered
dependency graph enriched for constituency recovery, nor a post hoc conversion
from a completed dependency graph. It is a constituency-style ordered tree
defined directly over the transition sequence.

The result also characterizes projectivity. A single-headed dependency tree
admits such a lexicalized ordered representation, with leaves in surface order
and contiguous subtree yields, if and only if it is projective. Projective
arc-standard derivations therefore do not merely accumulate dependency arcs;
they determine a recoverable ordered hierarchical object. For non-projective
inputs, the same interpretation can be used in practice through
pseudo-projective lifting before derivation and inverse decoding after recovery,
but the formal claim is specific to the projective case.

The contributions are threefold. First, we define an ordered tree interpretation
directly over arc-standard transition sequences. Second, we prove that the
representation is uniquely determined by the derivation, that the dependency
arcs are recoverable from it, and that its existence characterizes projectivity.
Third, we show in a small implementation study that the mapped derivations can
be executed in an existing neural transition-based parsing architecture.

\section{Background and related work}
\label{sec:related}

\paragraph{Transition systems and projectivity}
Arc-standard dependency parsing derives dependency trees through \textsc{shift}, \textsc{leftarc}, and \textsc{rightarc} transitions over a stack--buffer configuration \citep{yamada-matsumoto-2003-statistical,nivre-2003-efficient}. In its basic form, the system derives projective trees: attachments are created only when the relevant head and dependent are simultaneously available at the top of the stack. This restriction is central to the present paper, since the ordered representation defined here requires every derived subtree to have a contiguous surface yield.

Transition-based constituency parsers are likewise defined over stack--buffer computations, but their primitive objects are ordered tree fragments rather than dependency arcs \citep{sagae-lavie-2005-classifier,zhu-etal-2013-fast,dyer-etal-2016-recurrent,liu-zhang-2017-order,gonzalez-rodriguez-2019-faster}. The connection pursued in this paper is therefore not a general equivalence between dependency and constituency parsing. It concerns a shared structural condition: under projectivity, arc-standard dependency derivations and constituency derivations both manipulate hierarchical objects whose yields are contiguous substrings.

Non-projective inputs are handled only through pseudo-projective lifting \citep{kahane-etal-1998-pseudo-projectivity,nivre-nilsson-2005-pseudo}. Lifting projects a non-projective dependency tree into a projective one before derivation and restores the original dependency structure after recovery. It is a practical extension of the implementation, not part of the projectivity characterization proved in the main text.

\paragraph{Conversion, ordering, and derivational interpretation}
Much prior work relates dependency and constituency through completed structures. Dependency analyses have been converted into phrase-structure representations \citep{xia-palmer-2001-converting,kong-etal-2015-transforming}, and dependency trees have been encoded as bracketed, linearized, or ordered outputs for linearization, surface realization, and sequence-based prediction \citep{bohnet-etal-2012-generating,song-etal-2018-neural,puzikov-gurevych-2018-binlin,yu-etal-2020-fast}. These approaches show that dependency structures can be associated with tree-like or ordered representations, but the relevant object is normally introduced after the dependency graph has been constructed, or for a task external to the transition derivation itself.

The closest comparison is \citet{fernandez-gonzalez-martins-2015-parsing}, who introduce head-ordered dependency trees as an intermediate representation for reducing constituency parsing to dependency parsing. Their representation remains a dependency graph: head--dependent arcs are enriched with attachment-order information, and the resulting structure is parsed by a dependency parser. The ordering makes it possible to recover a headed constituency tree. \citet{morey-etal-2018-dependency} use this line of representation in RST discourse parsing, where headed discourse constituency structures are related to ordered dependency analyses.

The present paper differs in both direction and object type. It starts from dependency trees and arc-standard derivations, not from headed constituency treebanks. The induced object is a constituency-style ordered tree, not a dependency graph enriched with attachment order. Dependency arcs are recovered from this ordered tree, rather than constituency trees being recovered from an ordered dependency graph. The correspondence is therefore derivational rather than conversion-based: it identifies the hierarchical object constructed by an arc-standard transition sequence itself.

\section{Ordered tree interpretation of arc-standard derivations}
\label{sec:preliminaries}

We define the ordered tree object induced by arc-standard derivations and state its main formal properties. The result is stated for projective dependency trees; non-projective inputs are handled only as a practical extension through pseudo-projective lifting.

\subsection{Ordered dependency trees}

Let $w_{1:n}$ be a sentence in surface order, and let $w_0$ denote an artificial root. A dependency tree over $w_{0:n}$ is a directed, labeled tree $D=(V,E)$ with $V=\{0,\dots,n\}$, where each arc $(h,d,\ell)\in E$ specifies a head $h$, a dependent $d$, and a label $\ell$. Every token except $w_0$ has exactly one incoming arc.

For a projective dependency tree $D$, we define its ordered dependency tree representation, written $\mathcal{T}(D)$, as a rooted ordered tree whose leaves are the lexical tokens in surface order. Each internal node is anchored by a unique lexical head and combines that anchor with its dependent subtrees. Dependents to the left of the head appear to the left of the anchor, and dependents to the right appear to the right. The representation therefore preserves both lexical anchoring and surface-contiguous subtree yields. 


For completeness, Algorithm~\ref{alg:ud2tree_dlookup} gives the recursive construction of $\mathcal{T}(D)$ from a projective dependency tree. For each head $h$, the procedure visits $h$ and its dependents in surface order and returns a single ordered tree anchored at $h$.

\begin{algorithm}[!ht]
\caption{Deterministic construction of an ordered dependency tree representation}
\label{alg:ud2tree_dlookup}
\footnotesize
\begin{algorithmic}[1]
\STATE \textbf{Input:} dependency tree $D=(V,E)$ over $w_{0:n}$
\STATE \textbf{Output:} ordered tree $\mathcal{T}(D)$
\STATE Build adjacency lists $dlookup(h)=\{\,d \mid (h,d)\in E\,\}$ for all $h\in\{0,\dots,n\}$
\STATE Let $r$ be the unique dependent of $0$
\STATE \textbf{function} \textsc{Build}$(h)$
\begin{ALC@g}
  \STATE $C(h)\gets$ the elements of $\{h\}\cup dlookup(h)$ sorted increasingly
  \STATE $B\gets[\ ]$
  \STATE \textbf{for each} $i$ \textbf{in} $C(h)$ \textbf{do}
  \begin{ALC@g}
    \IF{$i=h$}
      \STATE $B.\textsc{append}(\textsc{Leaf}(w_h))$
    \ELSE
      \STATE $B.\textsc{append}(\textsc{Build}(i))$
    \ENDIF
  \end{ALC@g}
  \STATE \textbf{end for}
  \STATE \textbf{return} \textsc{Tree}($h,B$)
\end{ALC@g}
\STATE \textbf{end function}
\STATE \textbf{return} \textsc{Build}$(r)$
\end{algorithmic}
\end{algorithm}

Figure~\ref{ms-haag} illustrates the relation between a dependency graph and its ordered dependency tree representation.

\begin{figure*}[!ht]
\centering
\subfloat[Dependency graph\label{dependency-graph-book}]{
{\footnotesize
\begin{dependency}
  \begin{deptext}[column sep=0.5em]
\textit{book} \& \textit{me} \& \textit{the} \& \textit{morning} \& \textit{flight} \\
\textsc{vb} \& \textsc{prp} \& \textsc{dt} \& \textsc{nn} \& \textsc{nn} \\
1 \& 2 \& 3 \& 4 \& 5 \\
  \end{deptext}
  \deproot[edge below,edge unit distance=1.5ex]{1}{\textsc{root}}
  \depedge[]{1}{2}{\textsc{iobj}}
  \depedge[edge unit distance=2.5ex]{1}{5}{\textsc{dobj}}
  \depedge[]{5}{3}{\textsc{det}}
  \depedge[]{5}{4}{\textsc{compound}}
\end{dependency}
}
}
~~~
\subfloat[Ordered dependency tree representation\label{dependency-tree-book}]{
{\footnotesize
\branchheight{0.9cm}
\synttree
[\textsc{root}
    [\textit{book}/\textsc{vb}]
    [\textsc{iobj} [\textit{me}/\textsc{prp}]]
    [\textsc{dobj}
        [\textsc{det} [\textit{the}/\textsc{dt}]]
        [\textsc{compound} [\textit{morning}/\textsc{nn}]]
        [\textit{flight}/\textsc{nn}]
    ]
]
}}
\caption{A dependency graph and its ordered dependency tree representation. In the dependency graph, \textit{book} is the head of the \textsc{iobj} and \textsc{dobj} dependents, while \textit{flight} is the head of the \textsc{det} and \textsc{compound} dependents.}\label{ms-haag}
\end{figure*}

\begin{theorem}[Projective characterization]
\label{thm:projectivity}
A single-headed dependency tree admits a rooted ordered representation of the kind defined above, with leaves in surface order and every subtree spanning a contiguous substring, if and only if the dependency tree is projective.
\end{theorem}

\begin{proofsketch}
For a projective dependency tree, recursively expanding each head together with its dependents in surface order yields a rooted ordered tree whose subtree yields are contiguous. Conversely, if such an ordered representation exists, then no two dependency arcs can cross: crossing arcs would require the corresponding subtree yields to interleave, while subtree yields in an ordered tree with contiguity are necessarily disjoint or nested. Hence the dependency tree must be projective.
\end{proofsketch}

Thus the representation is not an arbitrary tree-shaped encoding of dependency structure. It is exactly the hierarchical object licensed by projectivity and surface contiguity.

\subsection{Arc-standard transitions as tree construction}

We now interpret arc-standard derivations as incremental construction of $\mathcal{T}(D)$. A mapped configuration is a triple $(\sigma,\beta,T)$, where $\sigma$ is a stack of partial ordered trees, $\beta$ is the unread buffer, and $T$ is the set of partial trees created so far. The action inventory remains the standard arc-standard inventory, but each transition is reinterpreted as a deterministic tree update.

A \textsc{shift} action introduces a singleton tree $t_i$ anchored at the next buffer token $w_i$ and pushes it onto the stack. A \textsc{leftarc} action applies to a stack $[\,\sigma \mid t_i \mid t_j\,]$ when $j \rightarrow i$; it attaches $t_i$ as a left dependent subtree of the head-anchored tree $t_j$ and returns the combined tree $t'_j$ to the stack. A \textsc{rightarc} action applies when $i \rightarrow j$; it attaches $t_j$ as a right dependent subtree of $t_i$ and returns the combined tree $t'_i$ to the stack. In both cases, the lexical anchor of the head subtree is preserved, and the dependent subtree is inserted on the side determined by surface order.

This gives a synchronous interpretation of arc-standard derivations. If $\theta$ is a dependency transition and $\hat{\theta}$ is its ordered-tree counterpart, then the mapping $f$ commutes with transition application:
\[
f\bigl(\theta(\sigma,\beta,A)\bigr)
=
\hat{\theta}\bigl(f(\sigma,\beta,A)\bigr).
\]
The mapped derivation maintains three invariants: each partial tree has a contiguous yield, each partial tree has a unique lexical anchor, and each attachment extends a head-anchored tree at the side licensed by projectivity.


The mapped transition system can be written explicitly as follows.
\paragraph{\textsc{shift}}
If the buffer has the form $[\,w_i,\beta\,]$, then \textsc{shift} creates a singleton tree $t_i$:
\begin{align*}
(\sigma,\,[\,w_i,\beta\,],\,T) &  \\
\longrightarrow  ([\sigma \mid t_i],\,\beta,\,T\cup\{t_i\}) & 
\end{align*}

\paragraph{\textsc{leftarc}}
If the stack has the form $[\,\sigma \mid t_i \mid t_j\,]$ and the dependency arc is $j\rightarrow i$, then:
\begin{align*}
([\sigma \mid t_i \mid t_j],\,\beta,\,T) 
&  \\
\longrightarrow   ([\sigma \mid t'_j],\,\beta,\,(T\setminus\{t_i,t_j\})\cup\{t'_j\})  &
\end{align*}
The new tree $t'_j$ preserves the anchor of $t_j$ and places $t_i$ as a left dependent subtree.

\paragraph{\textsc{rightarc}}
If the stack has the form $[\,\sigma \mid t_i \mid t_j\,]$ and the dependency arc is $i\rightarrow j$, then:
\begin{align*}
([\sigma \mid t_i \mid t_j],\,\beta,\,T)  & \\
 \longrightarrow  ([\sigma \mid t'_i],\,\beta,\,(T\setminus\{t_i,t_j\})\cup\{t'_i\}) & 
\end{align*}
The new tree $t'_i$ preserves the anchor of $t_i$ and places $t_j$ as a right dependent subtree.

\begin{theorem}[Transition correspondence]
\label{thm:transition-correspondence}
For every arc-standard derivation $\delta$ of a projective dependency tree $D$, the mapped transition sequence yields a well-defined ordered dependency tree representation $\mathcal{T}(\delta)$.
\end{theorem}

\begin{proofsketch}
The proof is by induction on derivation length. \textsc{shift} creates a singleton anchored tree. \textsc{leftarc} and \textsc{rightarc} combine the two topmost stack items by attaching one partial tree as a dependent of the other, while preserving the anchor of the head subtree. Projectivity guarantees that the dependent subtree is contiguous and lies on the appropriate side of the head anchor. Hence every mapped transition is well defined and preserves the invariants.
\end{proofsketch}

\begin{theorem}[Recoverability]
\label{thm:recoverability}
Let $\mathcal{T}(\delta)$ be the final ordered dependency tree representation produced by mapping an arc-standard derivation $\delta$ of a projective dependency tree $D$. Then the dependency arcs of $D$ are uniquely recoverable from $\mathcal{T}(\delta)$.
\end{theorem}

\begin{proofsketch}
Each internal combination preserves the lexical anchor of the head subtree and attaches one dependent subtree in an ordered position relative to that anchor. The head-bearing child identifies the head token, and each non-head dependent subtree contributes its lexical anchor as the dependent token. Traversing the ordered representation therefore reconstructs exactly the arcs introduced by the mapped derivation.
\end{proofsketch}

Theorems~\ref{thm:transition-correspondence} and~\ref{thm:recoverability} show that an arc-standard derivation does not merely construct dependency arcs. Under projectivity, it determines a recoverable ordered hierarchical object directly over the derivation.

\section{Executable derivations as tree parsing}
\label{sec:implementation}

The formal correspondence in Section~\ref{sec:preliminaries} defines an ordered tree object over arc-standard derivations. We use a small implementation study to verify that this object can function as a learnable parsing target, rather than to introduce a new parsing architecture.

We instantiate the mapped system in the \textsc{stanza} transition-based constituency parser.\footnote{\url{https://github.com/stanfordnlp/stanza/tree/main/stanza/models/constituency}} The encoder and training procedure are unchanged; the native constituency actions are replaced by the mapped actions \textsc{shift}, \textsc{leftarc}, and \textsc{rightarc}. The parser predicts the unlabeled ordered tree $\mathcal{T}(\phi(D))$, from which dependency arcs are recovered deterministically. Since dependency labels are not part of the mapped transition inventory, labeled scores are obtained by passing the recovered arcs to the \textsc{stanza} dependency labeling component. 
Appendix~\ref{sec:appendix-implementation} gives the implementation details.

Table~\ref{tab:main-results} reports dependency recovery from predicted ordered trees on PTB--WSJ converted to Stanford Dependencies \citep{de-marneffe-manning-2008-stanford}. The result shows that derivation-induced ordered trees provide stable targets for a constituency-parser framework while preserving the structural information needed for deterministic dependency recovery.

\begin{table}[!ht]
\centering
\footnotesize
\begin{tabular}{l cc}
\toprule
Dataset & UAS & LAS \\
\midrule
PTB--WSJ & 95.94 & 94.90 \\
\bottomrule
\end{tabular}
\caption{Dependency recovery from predicted ordered trees on PTB--WSJ.}
\label{tab:main-results}
\end{table}


\section{Discussion}
\label{sec:discussion}

Arc-standard projective derivations determine an ordered tree object, not merely a set of dependency arcs. The object is derivation-relative: it preserves lexical anchoring, maintains surface-contiguous subtree yields, and supports deterministic recovery of the arcs introduced by the derivation. The analysis therefore makes explicit a hierarchical structure already present in arc-standard parsing, but not usually stated as a tree-theoretic object.

This representation is not a general dependency-to-constituency conversion, nor does it imply that dependency and constituency structures are grammatically equivalent. It identifies a specific ordered hierarchy induced by arc-standard derivations under projectivity. The projectivity characterization is central: the representation is not merely compatible with projective dependency trees, but follows from the contiguity constraints imposed by projective arc-standard parsing. Non-projective dependencies cannot directly yield the same object without lifting, since crossing arcs would force interleaving subtree yields.

The comparison with head-ordered dependency trees \citep{fernandez-gonzalez-martins-2015-parsing} further clarifies the status of the representation. Head-ordered dependency trees remain ordered dependency graphs parsed by a dependency parser. The present object is instead a constituency-style ordered tree induced by dependency derivations and parsed as a tree object. The contrast is therefore representational as well as directional: dependency arcs are recovered from an ordered tree induced by the derivation, rather than constituency structure being recovered from an enriched dependency graph.

The ordered-tree interpretation allows dependency parsing to be recast as tree parsing. Since the induced object is linearly ordered, it can be used as the target structure of existing tree-based parsers while preserving deterministic dependency recovery. This recasting depends on the representation defined above: without a derivation-induced ordered object, an arc-standard dependency derivation would not provide a tree target with surface-ordered leaves and recoverable head--dependent relations. Using an existing in-order parser \citep{liu-zhang-2017-order}, this setting yields 96.76/95.92 UAS/LAS. A linearization parser \citep{vinyals-etal-2015-grammar} yields 96.68/95.44 UAS/LAS. These results do not define a new parsing architecture, but show that the induced object is computationally usable as a tree representation.

This use of linearization parsing differs from dependency tree linearization. In surface-realization settings, dependency linearization typically starts from an unordered dependency tree and predicts a surface order for its tokens \citep{puzikov-gurevych-2018-binlin,yu-etal-2020-fast}. BinLin constructs an auxiliary binary tree as an ordering device over dependency lemmas, while graph-based non-projective linearization formulates token ordering as a decoding problem over an unordered dependency input. In both cases, the dependency graph is given first, and linearization supplies the missing word order.

The present setting reverses this relation. The parser does not linearize an unordered dependency graph into a sentence; it predicts a linearized form of a derivation-induced ordered tree whose leaves are already constrained by surface order and whose internal structure is licensed by arc-standard computation. Dependency arcs are then recovered from the predicted tree object. Linearization is therefore not used to impose word order on a dependency graph, but to parse a surface-ordered tree object from which dependency structure is deterministically recoverable.

\section{Conclusion}
\label{sec:conclusion}

This paper has given a tree-theoretic interpretation of arc-standard projective dependency derivations. Under projectivity, the transition sequence induces a lexicalized ordered tree with surface-contiguous yields, and the dependency arcs introduced by the derivation are deterministically recoverable from that object. The correspondence is therefore derivational rather than conversion-based: the ordered hierarchy is not imposed after dependency parsing, but constructed by the arc-standard computation itself.

For non-projective inputs, the interpretation can be used through pseudo-projective lifting before derivation and inverse decoding after recovery. The formal characterization, however, remains specific to the projective case, where contiguity is preserved without additional encoding. More broadly, the analysis suggests that transition systems can be compared not only as procedures for constructing dependency graphs, but also by the hierarchical objects their derivations implicitly define.

\section*{Limitations}

The formal results are restricted to projective arc-standard derivations. This restriction is not an implementation artifact, but follows from the contiguity condition built into the ordered representation: non-projective arcs would require interleaving subtree yields and therefore cannot be represented directly by the same tree object. Pseudo-projective lifting provides a practical extension, but it does not change the theoretical characterization, which remains a statement about projective dependency trees.

The representation is also tied to arc-standard derivations. Other transition systems may induce different hierarchical objects, or none with the same recoverability and contiguity properties. The present analysis therefore should not be read as a general equivalence between dependency parsing and constituency parsing, but as a specific characterization of the ordered tree structure implicit in projective arc-standard parsing.


\appendix

\section{Full step-by-step transition example}
\label{sec:full-example}

This appendix gives a complete derivation for the example in
Figure~\ref{ms-haag}.\footnote{The sentence is adapted from
\citet[Chapter 19, p.\,9]{jurafsky-martin-2026-book}.}
The derivation shows how an arc-standard transition sequence incrementally
constructs the ordered dependency tree representation. Each \textsc{shift}
introduces a singleton anchored subtree; each \textsc{leftarc} or
\textsc{rightarc} combines the two topmost subtrees while preserving the lexical
anchor of the head.

\begin{table*}[!ht]
\centering
\resizebox{\textwidth}{!}{
\footnotesize
\begin{tabular}{c r l c c}
\toprule
Step & Stack & Buffer & Action & Relation added \\
\midrule
0 & [root] & [book, me, the, morning, flight] & \textsc{shift} & \\
1 & [root, $t_{\text{book}}$] & [me, the, morning, flight] & \textsc{shift} & \\
2 & [root, $t_{\text{book}}$, $t_{\text{me}}$] & [the, morning, flight] & \textsc{rightarc} & book $\rightarrow$ me \\
3 & [root, $t'_{\text{book}}$] & [the, morning, flight] & \textsc{shift} & \\
4 & [root, $t'_{\text{book}}$, $t_{\text{the}}$] & [morning, flight] & \textsc{shift} & \\
5 & [root, $t'_{\text{book}}$, $t_{\text{the}}$, $t_{\text{morning}}$] & [flight] & \textsc{shift} & \\
6 & [root, $t'_{\text{book}}$, $t_{\text{the}}$, $t_{\text{morning}}$, $t_{\text{flight}}$] & [] & \textsc{leftarc} & flight $\rightarrow$ morning \\
7 & [root, $t'_{\text{book}}$, $t_{\text{the}}$, $t'_{\text{flight}}$] & [] & \textsc{leftarc} & flight $\rightarrow$ the \\
8 & [root, $t'_{\text{book}}$, $t''_{\text{flight}}$] & [] & \textsc{rightarc} & book $\rightarrow$ flight \\
9 & [root, $t''_{\text{book}}$] & [] & \textsc{rightarc} & root $\rightarrow$ book \\
10 & [root] & [] & \textsc{done} & \\
\bottomrule
\end{tabular}
}
\caption{Arc-standard transition sequence and stack evolution for the running example.}
\label{tab:full-example}
\end{table*}

Figure~\ref{step-by-step} gives representative snapshots of the induced ordered
tree construction. The mapped transition system constructs only the unlabeled
ordered structure; dependency relation labels are not predicted during tree
construction. Accordingly, relation nodes are marked \textsc{ua}, for
unlabeled arc. Labels are assigned after structural recovery, as described in
Appendix~\ref{sec:appendix-implementation}.

\begin{figure}[!ht]
\centering
\subfloat[Step 0\label{step0}]{
\begin{tabular}[b]{p{0.5cm} p{0.5cm}}
$t_{\text{book}}$: &
\vspace{0pt}\scriptsize{
\begin{forest}
  [{\textsc{ua}} [book]]
\end{forest}
}
\end{tabular}
}
\quad
\subfloat[Step 2\label{step2}]{
\begin{tabular}[b]{p{0.5cm} p{1cm}}
$t'_{\text{book}}$: &
\vspace{0pt}\scriptsize{
\begin{forest}
  [{\textsc{ua}} [book] [\textsc{ua} [me]]]
\end{forest}
}
\end{tabular}
}
\quad
\subfloat[Step 6\label{step6}]{
\begin{tabular}[b]{p{0.5cm} p{1cm}}
$t'_{\text{flight}}$: &
\vspace{0pt}\scriptsize{
\begin{forest}
  [{\textsc{ua}} [\textsc{ua} [morning]] [flight]]
\end{forest}
}
\end{tabular}
}
\quad
\subfloat[Step 7\label{step7}]{
\begin{tabular}[b]{p{0.5cm} p{1.5cm}}
$t''_{\text{flight}}$: &
\vspace{0pt}\scriptsize{
\begin{forest}
  [{\textsc{ua}} [\textsc{ua} [the]] [\textsc{ua} [morning]] [flight]]
\end{forest}
}
\end{tabular}
}

\subfloat[Step 8\label{step8}]{
\begin{tabular}[b]{p{0.5cm} p{3cm}}
$t''_{\text{book}}$: &
\vspace{0pt}\scriptsize{
\begin{forest}
  [{\textsc{ua}}
      [book]
      [\textsc{ua} [me]]
      [\textsc{ua} [\textsc{ua} [the]] [\textsc{ua} [morning]] [flight]]
  ]
\end{forest}
}
\end{tabular}
}
\qquad
\subfloat[Step 9\label{step9}]{
\begin{tabular}[b]{p{0.5cm} p{3cm}}
$T_{\text{book}}$: &
\vspace{0pt}\scriptsize{
\begin{forest}
  [{\textsc{root}}
      [book]
      [\textsc{ua} [me]]
      [\textsc{ua} [\textsc{ua} [the]] [\textsc{ua} [morning]] [flight]]
  ]
\end{forest}
}
\end{tabular}
}

\caption{Representative snapshots of the ordered tree obtained after selected steps of the transition sequence in Table~\ref{tab:full-example}.}\label{step-by-step}
\end{figure}

\section{Technical realization}
    \label{sec:appendix-implementation}

The implementation uses the \textsc{stanza} constituency-transition framework. Its transition mechanism operates over a stack, a
buffer, and a sequence of discrete actions. We retain the neural encoder,
feature extraction, and training objective, and replace only the native
constituency action inventory with the dependency-motivated transitions
\textsc{shift}, \textsc{leftarc}, and \textsc{rightarc}. The resulting model
predicts the canonical transition sequence constructing
$\mathcal{T}(\phi(D))$.

\paragraph{Label prediction and inverse decoding}

The mapped transition system predicts the unlabeled ordered structure
$\mathcal{T}(\phi(D))$. Since dependency labels are not part of the mapped
transition inventory, labeled attachment scores are obtained by a separate
labeling step. We use the \textsc{stanza} graph-based dependency component
\citep{qi-etal-2020-stanza} for this purpose. In the standard graph-based
pipeline, the model determines head indices and labels the resulting arcs. In
our configuration, head selection is bypassed: the head--dependent relations are
fixed by the ordered tree recovered from the transition system, and the graph
component is used only to assign labels to those arcs.

For non-projective inputs, training and decoding are performed over the pseudo-projective tree $\phi(D)$. After label prediction, the enriched labels are decoded with $\phi^{-1}$ to recover the original dependency tree:
{\footnotesize
\begin{align*}
& D & \\
\;\xrightarrow{\;\phi\;}\; &
\phi(D) & \\
\;\xrightarrow{\;\textsc{mapped transition system}\;}\;
& \mathcal{T}(\phi(D)) & \\
\;\xrightarrow{\;\textsc{label prediction}\;}\;
&\mathcal{T}(\phi(D))^{\ell'} & \\
\;\xrightarrow{\;\phi^{-1}\;}\;
&D
\end{align*}}

The pipeline therefore separates three operations that are conceptually distinct:
projectivization, unlabeled ordered-tree prediction, and dependency-label
assignment. Only the middle step instantiates the proposed transition mapping;
pseudo-projective decoding and label prediction are external components used to
recover labeled dependency trees for evaluation.





\end{document}